\PassOptionsToPackage{table}{xcolor}
\documentclass[letterpaper, 10 pt, conference]{ieeeconf}  

\IEEEoverridecommandlockouts                              

\usepackage{times}
\usepackage{graphicx}
\usepackage{booktabs}
\usepackage{epsfig} 
\usepackage{amsmath} 
\usepackage{amssymb}  
\usepackage{breqn}
\usepackage{cite}
\usepackage{bm}
\usepackage{multicol, blindtext}
\usepackage{makecell}
\usepackage{float}
\usepackage{siunitx}
\usepackage{xcolor}
\usepackage{url}
\usepackage{hyperref}
\usepackage{lscape} 
\usepackage{multirow}
\usepackage{graphics}
\usepackage{pdfpages}

\DeclareGraphicsExtensions{.png,.pdf}

\title{\LARGE \bf

GraphGarment: Learning Garment Dynamics for Bimanual Cloth Manipulation Tasks

}

\author{~Wei~Chen,~Kelin~Li,~Dongmyoung~Lee,~Xiaoshuai~Chen,~Rui~Zong and~Petar~Kormushev
\thanks{~Wei~Chen,~Kelin~Li,~Dongmyoung~Lee,~Xiaoshuai~Chen,~Rui~Zong and~Petar~Kormushev are with the Dyson School of Design Engineering, Imperial College London, 25 Exhibition Road, London, SW7 2DB, UK
{\tt\small (w.chen21)@imperial.ac.uk}}
}
\begin{document}
\maketitle

\thispagestyle{empty}
\pagestyle{empty}

\begin{abstract}
Physical manipulation of garments is often crucial when performing fabric-related tasks, such as hanging garments. However, due to the deformable nature of fabrics, these operations remain a significant challenge for robots in household, healthcare, and industrial environments. In this paper, we propose GraphGarment, a novel approach that models garment dynamics based on robot control inputs and applies the learned dynamics model to facilitate garment manipulation tasks such as hanging. Specifically, we use graphs to represent the interactions between the robot end-effector and the garment. GraphGarment uses a graph neural network (GNN) to learn a dynamics model that can predict the next garment state given the current state and input action in simulation.
To address the substantial sim-to-real gap, we propose a residual model that compensates for garment state prediction errors, thereby improving real-world performance.
The garment dynamics model is then applied to a model-based action sampling strategy, where it is utilized to manipulate the garment to a reference pre-hanging configuration for garment-hanging tasks. 
We conducted four experiments using six types of garments to validate our approach in both simulation and real-world settings. 
In simulation experiments, GraphGarment achieves better garment state prediction performance, with a prediction error 0.46 cm lower than the best baseline. Our approach also demonstrates improved performance in the garment-hanging simulation experiment—with enhancements of 12\%, 24\%, and 10\%, respectively. Moreover, real-world robot experiments confirm the robustness of sim-to-real transfer, with an error increase of 0.17 cm compared to simulation results. Supplementary material is available at: \href{https://sites.google.com/view/graphgarment}{https://sites.google.com/view/graphgarment}.

\end{abstract}

\section{Introduction}


Manipulating garments is a key part of human daily routines. There is an increasing demand for robots to assist with tasks such as folding laundry, dressing, and hanging garments~\cite {9721534}. These tasks require the ability to perceive, manipulate, and plan to handle deformable objects, such as garments. While significant progress has been made in manipulating rigid objects~\cite{chen2025backbone,chi2023diffusion}, research in deformable object manipulation has largely focused on items with relatively simple topologies, such as ropes and square cloth. However, due to their complex dynamics, techniques for handling actual garments still remain in the early stages of development~\cite{9097275}.

%
Research involving cloth manipulation has been generally categorized into robotic system hardware design \cite{10410659, 9162474} and control algorithm design~\cite{10414107, 10342086, zhang2020learning,mo2022foldsformer}. 
Regarding the control aspect, perception and state estimation are critical factors in manipulating cloth-like objects. Due to their deformable nature, garments can adopt a wide range of configurations, making their manipulation more complex. Unlike regular square pieces of fabric, garments typically exhibit an enclosed structure, which adds further challenges when reasoning about their dynamics and action planning for manipulation. To address these challenges, a bimanual robot setup is often employed to constrain the garment deformation, thereby simplifying the perception during manipulation. However, while this bimanual setup helps manage deformation, it also introduces greater complexity in control and data acquisition. In this context, the representation of the garment state becomes crucial for understanding and controlling its dynamics.

\begin{figure}[t!]
    \centering
    \vspace{4mm}
    \includegraphics[width=0.98\columnwidth]{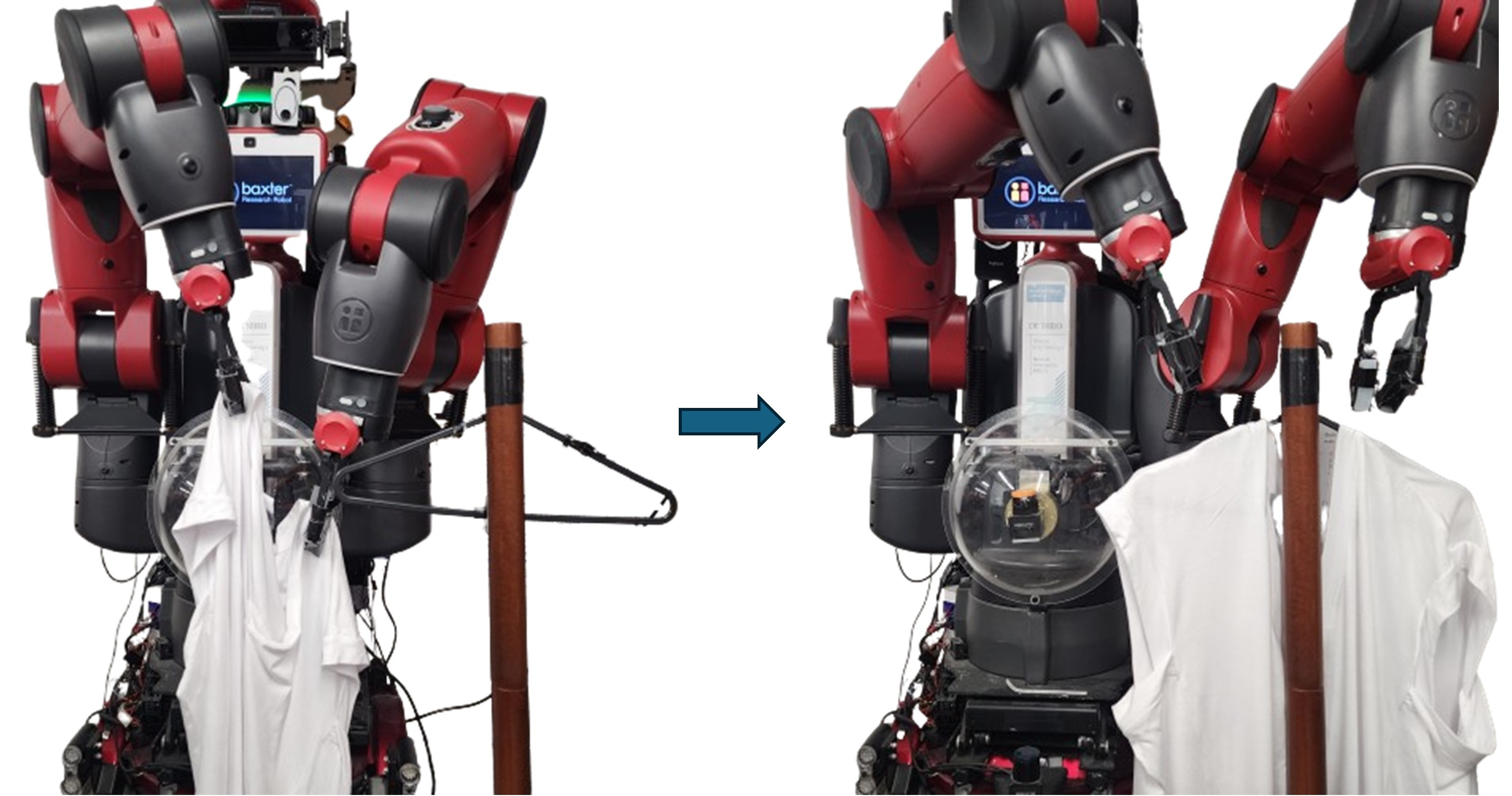}
    \caption{Our proposed approach, \textbf{GraphGarment}, allows a robot to learn a garment dynamics model that can predict the next state of the garment given the current garment state and robot action input. This is then used for the cloth manipulation of hanging the garment on a clothes hanger.}
    \label{fig:Intro}
    \vspace{-10pt}
\end{figure}


Existing research has explored various cloth representations to facilitate learning and control. With significant advancements in computer vision, studies have utilized image-based perception techniques, such as corner detection for cloth folding and collar detection for hanging garment~\cite{seita_bedmake_2019, 10342086}. However, due to issues such as self-occlusion and the deformable nature of cloth, inferring states directly from raw image inputs is challenging. 
Additionally, 3D representations such as point clouds have also been employed for garment state estimation~\cite{chi2021garmentnets, zhou2024bimanual}. 
Furthermore, recent research has investigated GNN-based approaches to enhance learning and control in this area~\cite{10611478}. However, these methods, including GNN and point cloud-related approaches, impose high demands on data acquisition to train the dynamics model. In addition, the model learned in simulation is difficult to apply directly to the real-world due to the large sim-to-real gap.

To learn the dynamics or perception of complex-shaped garment items, existing approaches typically require high-quality data in the real-world, which could be labour-intensive for data collection. 
In this paper, we propose \textbf{GraphGarment}: a novel approach that builds two models: a GNN model and a residual model to effectively learn garment dynamics in simulation and the real-world. Specifically, our approach first trains a GNN to predict the garment's next state based on the current state and action in simulation. With a small amount of real-world data, a residual model is then learned to bridge the sim-to-real gap. Leveraging the predictive model, a model-based action sampling strategy is proposed for the robot to adjust the garment to the optimal pre-hanging configuration. Finally, we utilize a novel robotic bimanual garment-hanging experiment to demonstrate the effectiveness of our method both in simulation and the real-world, as illustrated in Fig.\ref{fig:Intro}.



The key contributions of our proposed \textbf{GraphGarment} approach are: (1) a novel GNN-based model for learning the dynamics of garments; (2) a novel residual learning model to solve the sim-to-real challenges for garment manipulation; (3) a novel model-based action strategy for pre-hanging adjustment for the garment-hanging task.

\section{Related Work}
\subsection{Data-driven Approaches for Cloth Manipulation}

Perception and action planning are fundamental components in the manipulation of objects, particularly when dealing with deformable objects like clothes. Due to its inherent deformation characteristics and high elasticity, cloth presents significant non-linearity, making accurate state estimation challenging. This complexity can result in unpredictable behaviour during manipulation, necessitating sophisticated approaches to ensure precise control. Recent progress in learning-based methods has led to the emergence of various data-driven approaches, which have proven effective in addressing these challenges.

Different representations can be used for learning cloth manipulation. For instance, they can be modelled as 1-D objects (such as rope segments), 2-D representations (like images of garments), or 3-D structures (including point clouds for detailed shape and pose estimation). In the context of these representations, research efforts have explored several innovative solutions. Deep learning methods for 1-D rope perception have enabled more accurate tracking of linear and flexible objects\cite{zhaole2023robust}, while 2-D structural detection techniques have enhanced the identification of key regions in garments, such as collar\cite{10342086}. Additionally, 3-D garment pose estimation using point clouds has provided an understanding of the spatial arrangement and shape of cloth\cite{chi2021garmentnets}.

Data-driven control methods such as using reinforcement learning (RL) and imitation learning (IL) for decision-making\cite{10414107}, perception for manipulation and grasping\cite{10342086}, learning-based model predictive control for human-robot interaction\cite{10436357} have also made notable progress in data-driven control within the field of cloth manipulation.

\begin{figure*}[t!]
    \centering
    \vspace{4mm}
    \includegraphics[width=1.98\columnwidth]{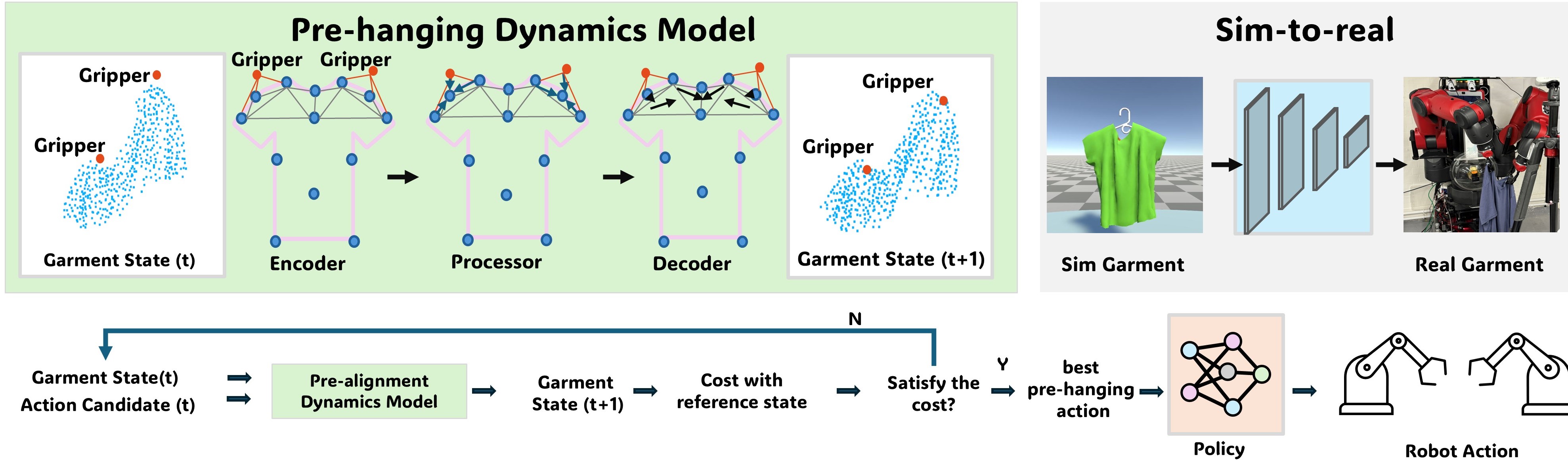}
    \caption{This figure demonstrates the overview of \textbf{GraphGarment}: A GNN predicts garment dynamics in simulation, refined by a residual model with real-world data. Using this predictive model, a model-based action sampling strategy adjusts the garment for optimal pre-hanging. A bimanual robotic hanging experiment validates the approach in both simulation and reality. }
    \label{fig:Main_fig}
    \vspace{-10pt}
\end{figure*}

\subsection{GNNs for Cloth Dynamics Learning}
Recent studies have successfully applied GNNs to learn the dynamics of complex simulated systems~\cite{sanchez2020learning}. For instance, \cite{pfaff2021learning} demonstrated a mesh-based method that effectively learns the dynamics of physical systems in a purely simulated environment. Although these methods perform strongly in simulation, effectively deploying them in the real-world scenarios with robotics-related applications remains a major challenge due to the significant sim-to-real gap. Several studies have been conducted in using GNNs for model-based manipulation tasks. Despite the sim-to-real gap, \cite{tuomainen2022manipulation} applied GNNs to learning the particle interaction for the manipulation of granular material. Other researchers have attempted to gather data directly from real-world scenarios—for instance, \cite{zhou2024bimanual} labels the rim area of garments and uses a GNN-based network to predict its position in subsequent time steps. However, this approach requires additional labeling of target areas and incurs high data collection costs in real-world settings. Therefore, how to effectively transfer models trained in simulation to real-world applications remains a significant challenge.

\subsection{Garment-hanging Task}
Recent research on garment-hanging tasks emphasizes using data-driven approaches, particularly deep learning, to tackle garment manipulation problems. Early work has been done by using RL to pick up the squared cloth and place it over a hanger~\cite{matas2018sim}. Additionally, Some studies relied on a peg-based cloth tree structure for garment-hanging, which simplifies the control requirements~\cite{10342086, wu2024unigarmentmanip}. With these methods, the system’s main challenge is identifying a suitable grasping point, as the peg-based structure simplifies the overall control needed for hanging the garment. 

In some domestic scenarios, garments are typically hung on triangular hangers. To achieve such hanging tasks, the system needs not only advanced perception but also sophisticated control methods to manage the complex interactions among the hanger, the garment, and the end effector. This necessitates a more comprehensive approach to accurately predict the garment’s dynamics and ensure precise manipulation during the hanging process. In this paper, we focus on modelling these complex garment dynamics and applying the learned model to guide decision-making within the overall garment manipulation system.

\section{Methodology}
\begin{figure}[t!]
    \centering
    \includegraphics[width=0.99\columnwidth]{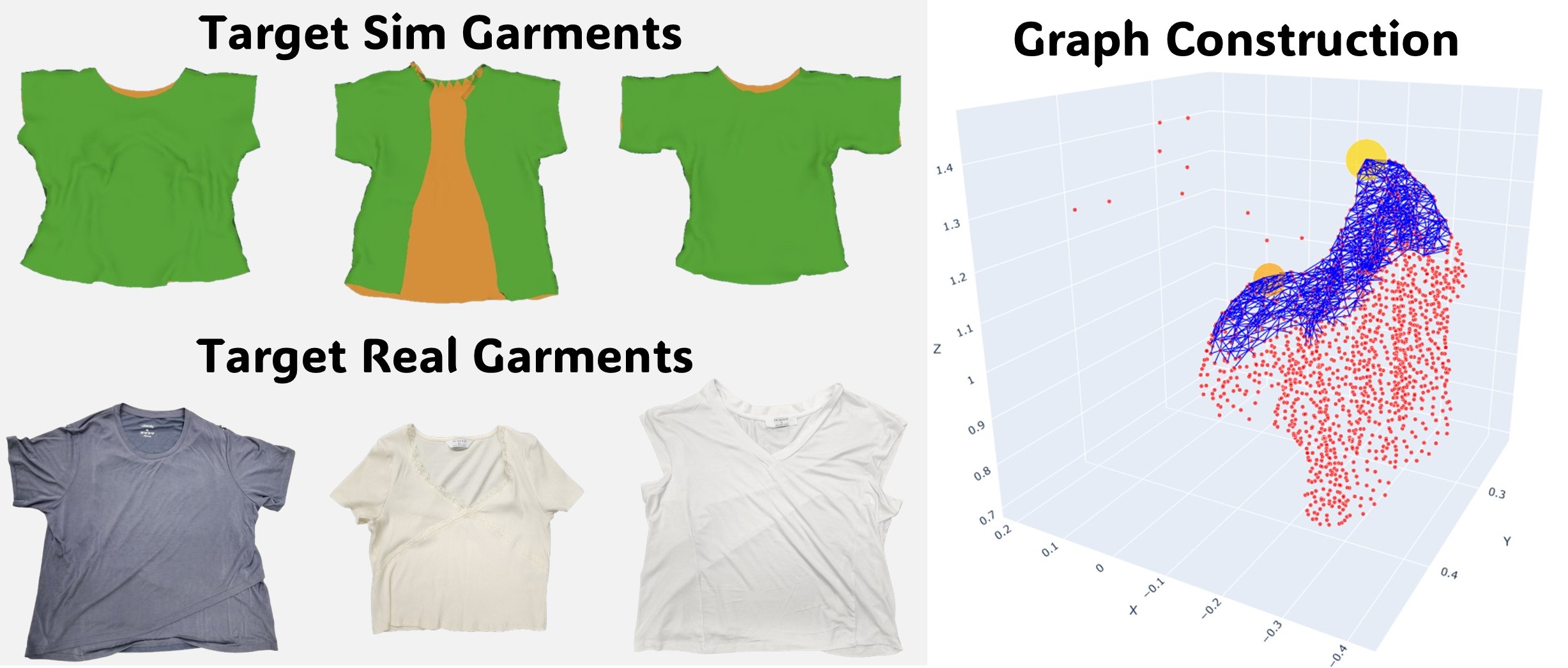}
    \caption{Left: The left figure shows the experiment's target garments, including simulated (Garment 1, Garment 2, Garment 3) and real garment (Garment 1, Garment 2, Garment 3). Right: The right figure displays a sample graph illustrating its construction.}
    \label{fig:Method_1}
    \vspace{-10pt}
\end{figure}

\subsection{Garment Dynamics Learning}

\subsubsection{Data Acquisition}
A major challenge in garment manipulation is collecting training data, which involves both simulating garments and gathering expert demonstrations. We address this by using the Unity simulation environment with the Obi Cloth Physics engine. Our research tackles two tasks: learning garment dynamics and performing bimanual garment-hanging.

For the garment dynamics task, we sample pre-hanging actions from a normal distribution and record $\{(s_t,a_t), s_{t+1}\}$ transition pairs to train the dynamics model. For the expert demonstrations, we build a VR-based system (inspired by \cite{10203548}) and use an HTC VIVE controller for pick-and-place actions, allowing us to directly capture human expert trajectories rather than relying on hand-crafted policies.

\subsubsection{Graph Representation}
In this step, we convert the garment point cloud into a graph representation. Specifically, the garment graph is defined as $G = (V, E)$, where $V=\{v_i\}_{i=1}^{N}$ represents the vertices of garments, with $v_i=(x_i, c_j)_{j=1}^{4}$, denoting the 3D positions and types of garment vertices. Within each graph, the vertices are categorized into four types: \textbf{garment\_left\_grasped}, \textbf{garment\_right\_grasped}, \textbf{main body}, and \textbf{action nodes}. Specifically, we sample 250 points around the left and right action nodes to obtain the left grasped area and right grasped area, respectively, with the remaining points forming the garment’s main body. $E$ denotes the edges that connect garment vertices. In our research, we adopt a K-nearest neighbour (KNN) scheme to determine the connections among vertices. A sample graph is illustrated in Fig.\ref{fig:Method_1}.
In addition, for most garment-related tasks, the primary focus is on the area currently held by the robot gripper. Therefore, we only construct edges among \textbf{garment\_left\_grasped}, \textbf{garment\_right\_grasped}, and \textbf{action nodes}. With this design choice, the information is only updated at these points, enabling us to predict the grasped area in the next timestep without considering the unnecessary sections of the garment’s main body, which can be found in the Fig.\ref{fig:Main_fig}.

\subsubsection{Forward Graph Neural Network}
The objective of the forward GNN model is to predict the 3D positions of garment particles in the next time step, given the current state and action.
$$y_{t+1} = f_{dyn}(s_{t}, a_t)$$
While a complete prediction is not necessary, here we only want to predict the area of \textbf{garment\_left\_grasped}, \textbf{garment\_right\_grasped} for manipulation tasks. We construct our GNN model $f_{dyn}$ using an Encoder-Processor-Decoder architecture and train it in a supervised manner following~\cite{pfaff2020learning}.  

\textbf{Encoder}: Once the graph is constructed, the input will be first encoded to a latent space with a feature size of 128 for each vertex feature and edge feature. We use multilayer perceptions (MLP) to build our encoders.
$$\hat{v} = \phi_{vert}(v) \qquad \hat{e} = \phi_{edge}(e) $$

\textbf{Processor}: Several identical message-passing blocks are then applied to update the latent graph from vertices and edges based on their connectivity. Specifically, it is defined as follows:
$$v_{i} = \phi_{vert}(v_{i},\sum_{j}{e_{ij}}) \qquad e_{ij} = \phi_{edge}(e_{ij}, v_{i}, v_{j},) $$
where each block here is implemented using MLPs with residual connections.

\textbf{Decoder}: In order to obtain the predicted state of the next timestep $t+1$, we need to apply another decoder to transform the graph from the latent feature space to the final prediction:
$$v_{predict} = \delta_{decoder}(v)  $$

We treat our system as quasi-static and aim to predict the 3D position of the garment's grasped area in the next state. Since we can easily access the indices of the garment points in the simulation environment, we use a mean squared error (MSE) loss between the predicted and ground truth garment positions to supervise the training of the GNN.

\subsection{Residual Model for Sim-to-real Transfer}
\begin{figure}[t!]
    \centering
    \vspace{4mm}
    \includegraphics[width=0.98\columnwidth]{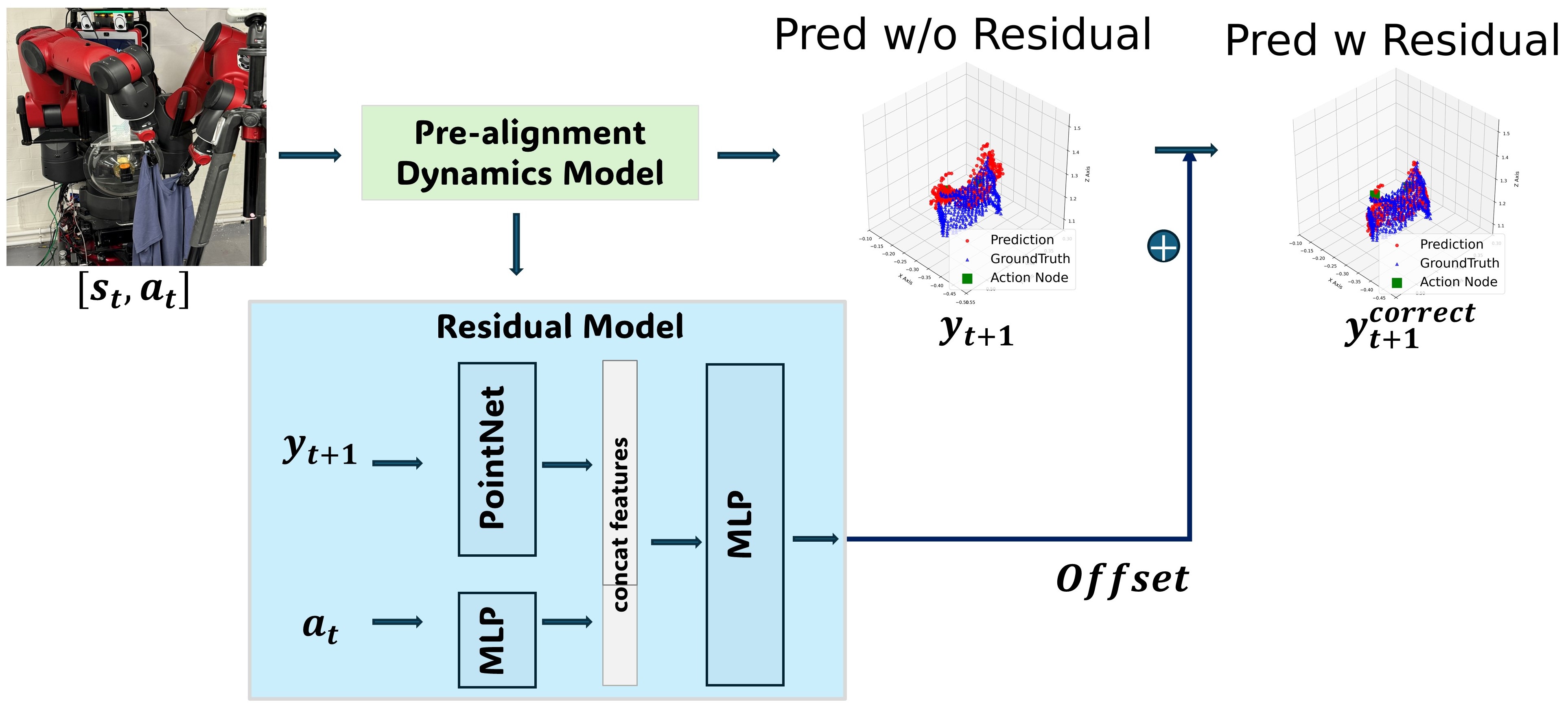}
    \caption{Residual Network: A PointNet-based residual network is utilized here to generate an offset value for each predicted point to correct the output of the dynamics model.}
    \label{fig:Method_2}
    \vspace{-10pt}
\end{figure}

Directly applying a garment dynamics model learned from simulation to the real-world presents a significant challenge due to the sim-to-real gap. This issue is twofold: variations in material properties and cloth size cause real-world cloth dynamics to differ from those in simulation. Accurate state estimation of the garment is difficult due to environmental noise and self-occlusion. To mitigate this issue, we learn an extra residual model to correct the predicted point cloud in real-world. The model takes the current action and prediction of the GNN model to output an offset value for each point of the predicted point cloud as illustrated in Fig.\ref{fig:Method_2}. Instead of fine-tuning the entire GNN model, we aim to apply this offset value to refine local predictions, effectively bridging the sim-to-real gap. Specifically, we use the PointNet-based~\cite{qi2017pointnet} architecture for the training of the residual model. It processes the point cloud to extract local and global features, embeds the action, and predicts offsets to correct the point cloud. The loss is computed by calculating the Chamfer distance between the ground truth and the corrected prediction.
\begin{equation*}
    \begin{split}
        y^{correct}_{t+1} &= f_{res}(y_{t+1}, a_{t}) + y_{t+1}   \\
        Loss_{res} &= \textit{chamfer\_dist}(s_{t+1}, y^{correct}_{t+1})
    \end{split}
\end{equation*}

\subsection{Garment Pre-hanging Adjustment for Hanging}
\subsubsection{Model-based Action Sampling}
Handling garment deformation is one of the most critical tasks in garment manipulation. In this section, we propose a model-based action sampling strategy for the garment pre-hanging adjustment. Before hanging the garment, we first sample action candidates and use the dynamics model to predict the next state until the cost converges. 
To select the best grasping candidate, we use the Chamfer distance between the predicted garment state and a reference garment pre-hanging state as the cost function.
The action with the lowest cost is then used for adjusting the garment to an ideal configuration for the hanging trajectories as shown in Fig.\ref{fig:Main_fig}.


\subsubsection{Learning from Demonstration for Hanging}

Due to the deformable nature of garments and its typically high-dimensional configuration space, learning an end-to-end policy that can effectively manipulate garments remains a significant challenge. However, by using the pre-hanging dynamics model, the garment can be adjusted to a relatively optimal configuration. We therefore apply Learning-from-Demonstration (LfD) to replicate expert behaviours. Once we have this demonstration data, we train a Dynamic Movement Primitive (DMP) \cite{ijspeert2013dynamical} that captures the core motion dynamics. This DMP can then be adapted to different initial and goal states, enabling the robot to reproduce the learned motion for hanging garments, as shown in Fig.\ref{fig:Main_fig}.

\section{Experiments}

\begin{figure*}[t!]
    \centering
    \vspace{3mm}
    \includegraphics[width=1.95\columnwidth]{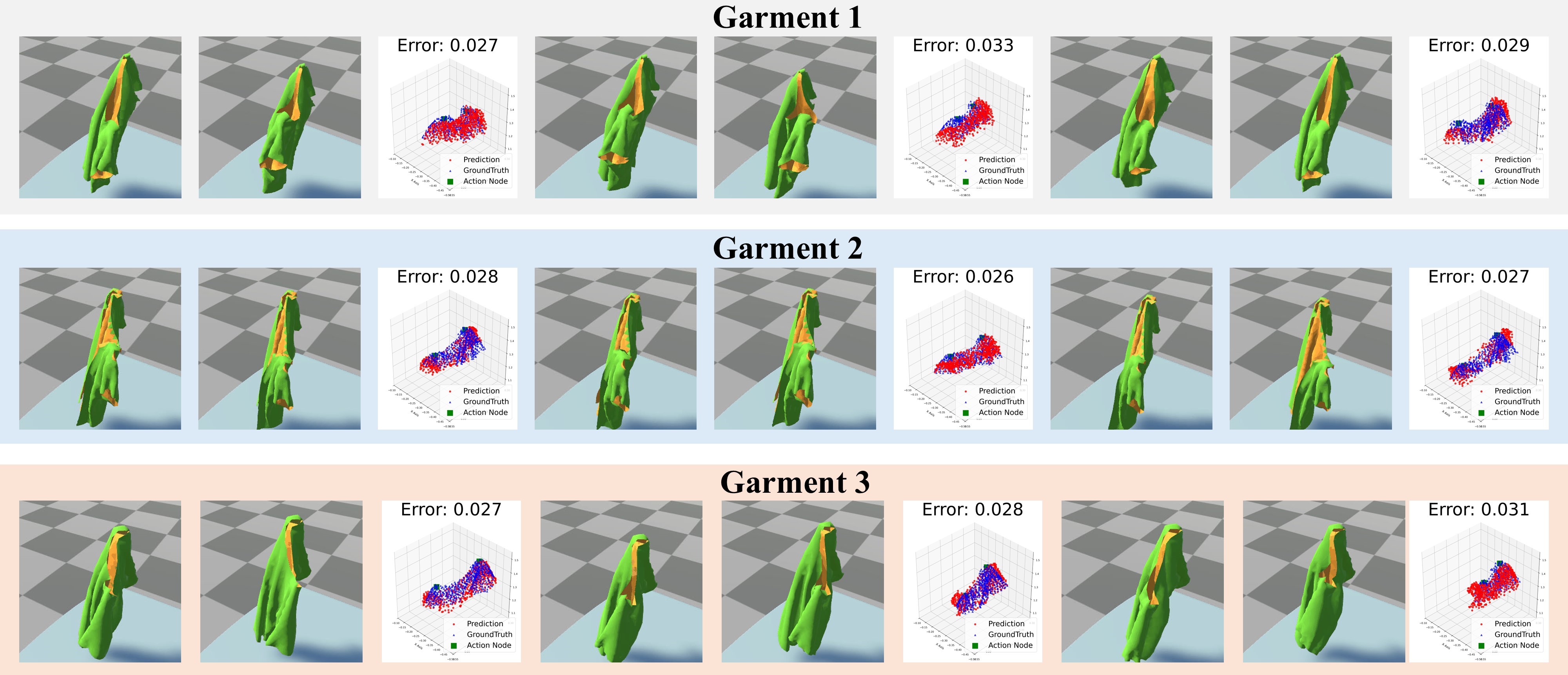}
    \caption{This figure shows our proposed approach for predicting the next state of the garment. We demonstrate the garment states at timesteps $s_t$ and $s_{t+1}$. We also present the prediction result, along with the Chamfer distance computed between the predicted state $y_{t+1}$. and the actual state $s_{t+1}$. Additional demonstrations are available in the multimedia resource.}
    \label{fig:Sim_demo1}
    \vspace{-10pt}
\end{figure*}

\begin{figure}[t!]
    \centering
    
    \includegraphics[width=0.98\columnwidth]{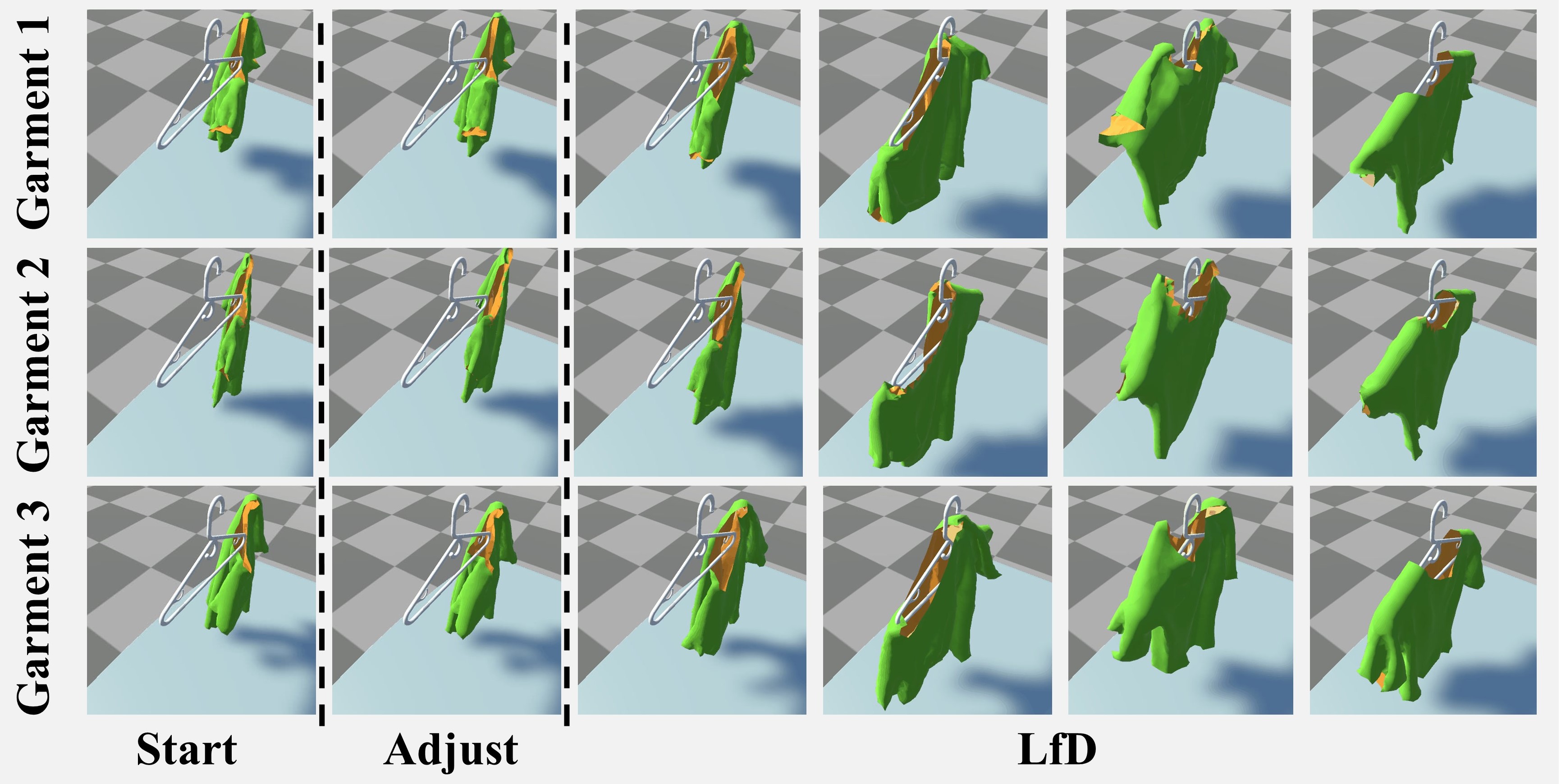}
    \caption{This figure illustrates our proposed approach for garment-hanging experiments. Additional demonstrations are available in the multimedia resource.}
    \label{fig:Sim_demo2}
    \vspace{-10pt}
\end{figure}

Simulation and real-world experiments are conducted to test the proposed garment manipulation system, with the target garments shown in Fig.~\ref{fig:Method_1}.
Specifically, we evaluate the proposed garment manipulation system in three aspects: 1) the accuracy of the predictive model for garment next state prediction against several baselines; 2) our modular design with end-to-end learning methods for garment-hanging tasks; 3) real-world experiment for the validation of proposed methods.

\subsection{Tasks and Implementation Details}

The implementation of our neural network is based on PyTorch. Specifically, we apply 15 message-passing layers to construct the GNN. 
For the data collection of garment dynamics learning, we collect 10,000 transition pairs of $\{(s_t,a_t), s_{t+1}\}$ for each garment.
For training our GNN, we apply Adam optimizer with a learning rate of 1e-4. A learning rate scheduler of gamma 0.5 is applied here with every 20 epochs. We train our model for 100 epochs until the training process converges.
All training is conducted on a machine equipped with an Intel Core i9 CPU and an Nvidia GeForce GTX 3080Ti GPU.

\subsection{Comparison Experiment On Garment State Prediction in Simulation}

In this experiment, we evaluate the accuracy of the garment dynamics model in simulation. Generally, garments can be represented as point clouds and graphs. Accordingly, we implement several baseline approaches that use these graph and point cloud representations as input to predict the next state of the garment. Specifically, the garment is initialized by randomly grasping a point around the collar and lifting it to the desired pre-hanging position. Following by introducing Gaussian noise to the pre-hanging position, we randomly sample actions, move the garment accordingly, and predict its next state based on the current state and the sampled action. We perform 100 trials for each type of garment. The chamfer distance between the prediction and the observed next garment state is used for the evaluation.

\begin{itemize}
\item
\textbf{Ours}: We use our proposed method to predict the next state of the garment in 3-D positions. The input is a KNN-based graph constructed from the current garment state and the gripper action.
\end{itemize}

\begin{itemize}
\item
\textbf{GCN-Dynamics}: We use graph convolutional neural networks~\cite{kipf2017semisupervised} to predict the next garment state in 3-D positions. The input is a KNN-based graph constructed from the current garment state and the gripper action.
\end{itemize}

\begin{itemize}
\item
\textbf{PointNet-Dynamics}: We use a PointNet encoder to extract a latent representation from the current garment point cloud, which is then concatenated with the gripper action node. A decoder then reconstructs the garment’s point cloud at the next time step from this combined representation.
\end{itemize}

\begin{itemize}
\item
\textbf{MLP-Dynamics}: We use an MLP to predict the next garment state point cloud. The input is the current garment point cloud concatenated with the gripper action node. A three-layer MLP is applied here for the feature extraction.
\end{itemize}

\begin{table}[t!]
\centering

\caption{Dynamics Model Simulation Results}
\vspace{-2mm}

\setlength{\tabcolsep}{1.7mm}{
\begin{tabular}{c|c c c | c}
\specialrule{1.3pt}{3pt}{2pt}
\textbf{Method / Tasks}  &\textbf{Garment 1} & \textbf{Garment 2}& \textbf{Garment 3} & \textbf{Average} \\ [0.2ex]
\hline
\makecell[c]{B1: GCN}       &3.77 &3.45 &3.87 & 3.69\\
\hline

 \makecell[c]{B2: PointNet}       &4.33&4.74 &4.57 &4.54\\
\hline
\makecell[c]{B3: MLP} &3.28&3.53 &3.52& 3.44 \\
\hline

 \rowcolor{gray!20}{\textbf{GraphGarment}}    &\textbf{2.94}&\textbf{2.84} &\textbf{3.16} &\textbf{2.98} \\ 
\specialrule{1.3pt}{1pt}{1pt}
\end{tabular}
}
\par
\begin{flushleft}
Chamfer distance (cm) between the predicted point cloud and the observed next-state point cloud is applied for evaluation. A smaller Chamfer distance indicates higher accuracy.

\end{flushleft}
\label{table:grasping}
\vspace{-10pt}
\end{table}

\begin{table}[t!]
\centering

\caption{garment-hanging Simulation Results}
\vspace{-2mm}

\setlength{\tabcolsep}{3.2mm}{
\begin{tabular}{c|c c c}
\specialrule{1.3pt}{3pt}{2pt}
\textbf{Method / Tasks}  &\textbf{Garment 1} & \textbf{Garment 2}& \textbf{Garment 3} \\ [0.2ex]

\hline
\makecell[c]{BC-PointNet}       &60\%&62\% &50\% \\
\hline
\makecell[c]{BC-MLP} &80\%&42\% &76\% \\

\hline

 \rowcolor{gray!20}{\textbf{GraphGarment}}    &\textbf{92\%}&\textbf{86\%} &\textbf{86\%}\\ 
\specialrule{1.3pt}{1pt}{1pt}
\end{tabular}
}
\par
\begin{flushleft}
Hanging experiments result in the simulation environment. Fifty trials are performed for each garment. 
\end{flushleft}
\label{table:grasping}
\vspace{-10pt}
\end{table}

\begin{figure}[t!]
    \centering
    \includegraphics[width=0.98\columnwidth]{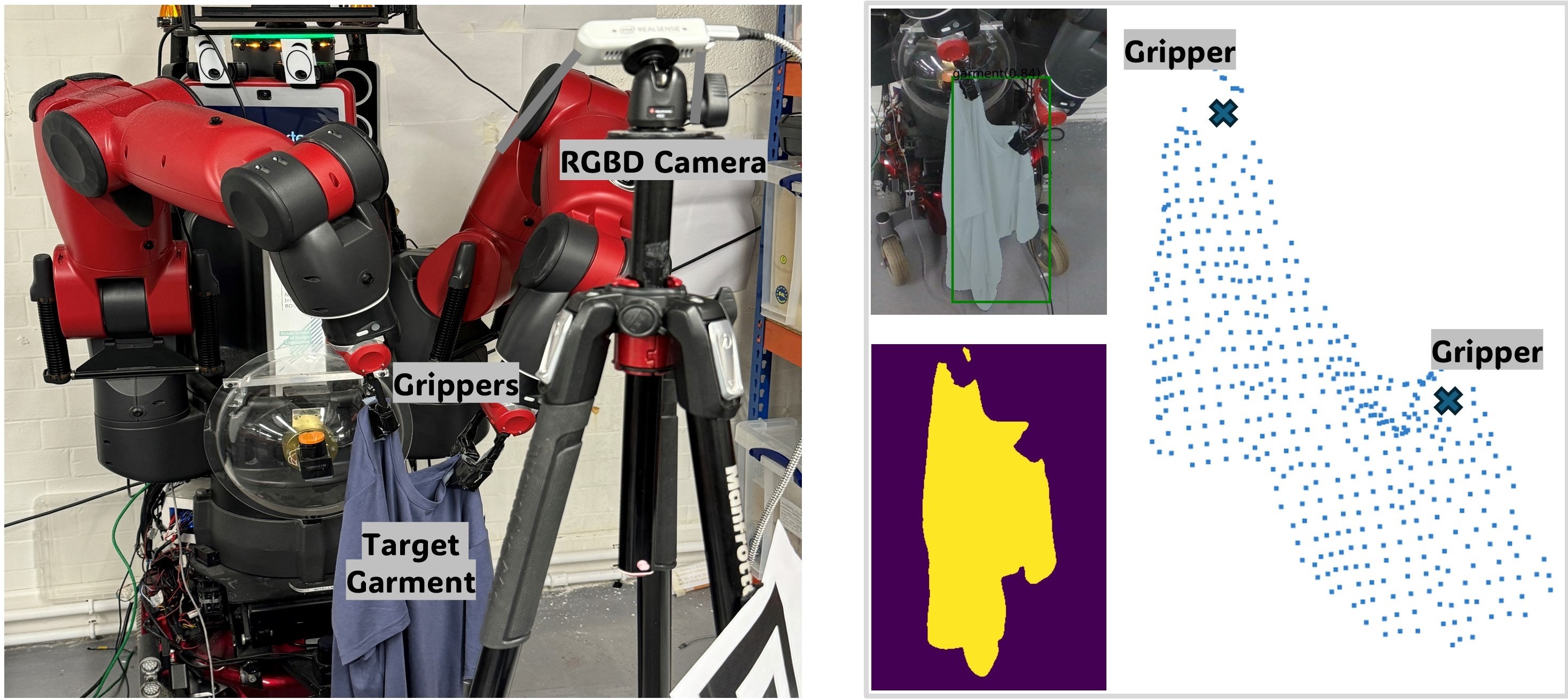}
    \caption{Real-world setup for cloth garment dynamics prediction task: A Baxter robot is used for garment manipulation. The results of Grounded SAM and the projected point cloud are also demonstrated here.}
    \label{fig:Experiment}
    \vspace{-10pt}
\end{figure}

\begin{figure*}[!t]
    \vspace{2.5mm}
    \centering
    \includegraphics[width=1.98\columnwidth]{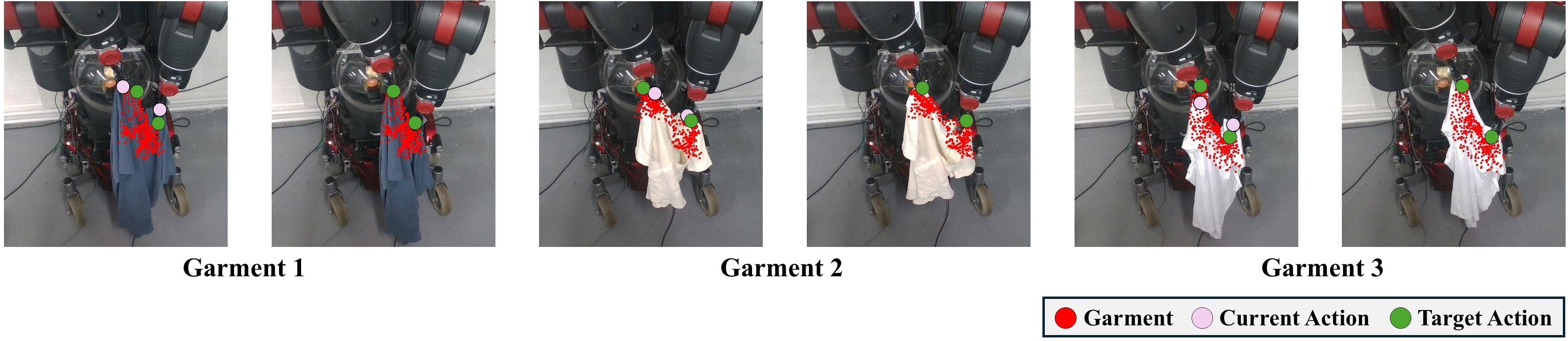}
    \caption{This figure illustrates our proposed approach for garment dynamics prediction experiments in the real-world. 
    The red dots correspond to the current garment state.
    The purple dots correspond to the current action. The green dots indicate the target action applied. Additional demonstrations are available in the multimedia resource.}
    \label{fig:Real_demo1}
    \vspace{-15pt}
\end{figure*}

\subsection{Comparison Experiment On Garment-hanging Success}
In this experiment, we evaluate the performance of the \textbf{GraphGarment} for garment-hanging tasks. We randomize the garment in a different configuration and grasp, to test the robustness. A trial is considered successful if both sleeves remain stable on each half of the cloth hanger. We perform 50 trials for each garment in the hanging experiment to obtain a robust result. For the baseline experiment, we implement two end-to-end imitation learning-based methods that use a garment point cloud as input and generate an action as the output.

\begin{itemize}
\item
\textbf{Ours}: We apply our complete method with garment pre-hanging adjustment and LfD for garment-hanging tasks.

\item
\textbf{BC-PointNet}: We apply a behaviour cloning policy that leverages point cloud and action pairs for policy training. To enhance performance, we incorporate additional proprioception information, specifically the end-effector position from the previous timestep, during model training and inference. The PointNet architecture is utilized to extract features from the observed garments.

\item
\textbf{BC-MLP}: We apply a behaviour cloning policy using point cloud and action pairs to replicate expert behaviours. Proprioception information is also applied here during training and inference. A three-layer MLP architecture is used to extract features from the observed garments.

\end{itemize}

\subsection{Real-world Experiment}
We evaluate our proposed methods in the real-world setting. Both the predictive model and hanging algorithm are validated here. We use the same evaluation method as mentioned in the simulation experiment section. To train the residual model, we follow the same data collection protocol described in Section III-A (1). In this process, we use the observed point cloud at the next step as the ground truth, while the input consists of the garment dynamics model's prediction and the current action.

\begin{figure}[t!]
    \centering
    
    \includegraphics[width=0.95\columnwidth]{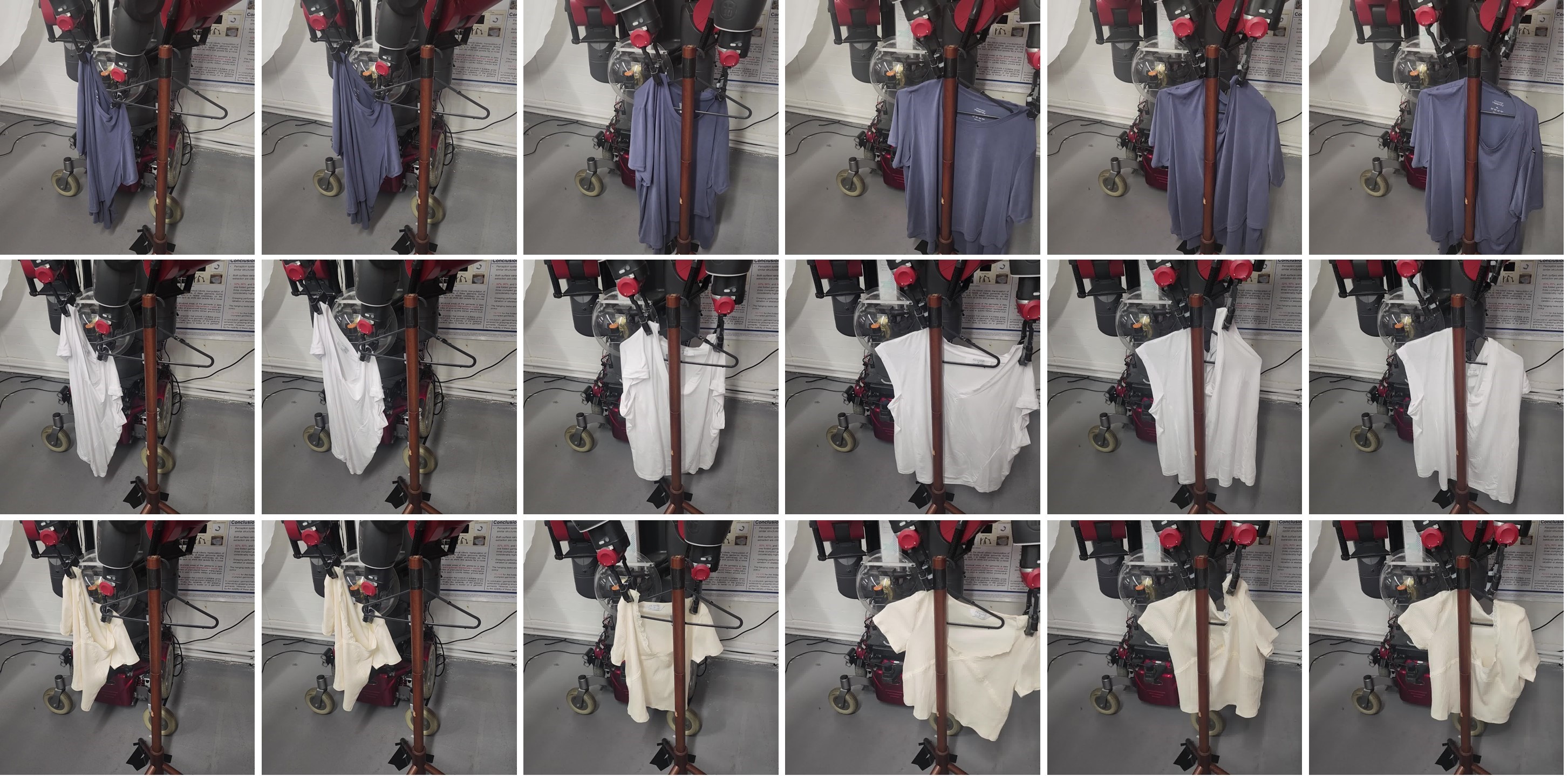}
    \caption{This figure illustrates our proposed approach for garment-hanging experiment in the real-world. Additional demonstrations are available in the multimedia resource.}
    \label{fig:Real_demo2}
    \vspace{-10pt}
\end{figure}

\begin{table}[t!]
\centering

\caption{Dynamics Model Real-world Results}

\setlength{\tabcolsep}{1.3mm}{
\begin{tabular}{c|c c c | c}
\specialrule{1.3pt}{3pt}{2pt}
\textbf{Method / Tasks}  &\textbf{Garment 1} & \textbf{Garment 2}& \textbf{Garment 3} & \textbf{Average} \\ [0.2ex]
\hline

{\textbf{Ours w/o Residual}}    &3.58&4.77 & 4.17 & 4.17 \\ 
\hline

{\textbf{Ours}}    &\textbf{3.11}&\textbf{3.15} &\textbf{3.18} &\textbf{3.15} \\ 
\specialrule{1.3pt}{1pt}{1pt}
\end{tabular}
}
\par
\begin{flushleft}
Chamfer distance (cm) between the predicted point cloud and the observed next-state point cloud is applied for evaluation. A smaller Chamfer distance indicates higher accuracy.

\end{flushleft}
\label{table:grasping}
\vspace{-10pt}
\end{table}

\subsubsection{Experiment Setting}
 We use a Baxter robot for the real-world experiments. With the garment grasped at two points by the robot's grippers, the target garment can be captured by an Intel RealSense 435i RGB-D camera. To obtain the garment point cloud, we first use Grounded SAM~\cite{ren2024grounded} with the text prompt "garment" input to segment garment pixels in 2D images. By aligning the RGB images and depth images, we can then project the garment's pixel into 3D space with the camera's intrinsic. The hand-eye calibration was also set up before the experiments. More detailed settings can be found in Fig.\ref{fig:Experiment}.

\subsubsection{Real-world Garment State Prediction Experiment}
We evaluate the accuracy of the predictive model in the real-world setting. Specifically, we ablate the residual model of \textbf{GraphGarment} to demonstrate its performance. Some sample runs are illustrated in Fig.\ref{fig:Real_demo1}. We project the 3D point prediction to the 2D images for better visualization. Results are presented in Table III.

\begin{table}[t!]
\centering

\caption{garment-hanging Real-world Results}
\vspace{-2mm}

\setlength{\tabcolsep}{3mm}{
\begin{tabular}{c|c c c}
\specialrule{1.3pt}{3pt}{2pt}
\textbf{Method / Tasks}  &\textbf{Garment 1} & \textbf{Garment 2}& \textbf{Garment 3} \\ [0.2ex]
\hline

{\textbf{Ours}}    &{26/30}&{23/30} &{25/30} \\ 
\specialrule{1.3pt}{1pt}{1pt}
\end{tabular}
}
\par
\begin{flushleft}

We carried out 90 trials overall to validate our system in real-world scenarios.

\end{flushleft}
\label{table:grasping}
\vspace{-10pt}
\end{table}

\subsubsection{Real-world Garment-hanging Experiment}

We evaluate the ability of overall capability of our proposed system based on a dual-arm garment-hanging task.  some samples are ilustrated in Fig.\ref{fig:Real_demo2}. Results are presented in Table IV.

\subsection{Discussion}

As shown in Table I, our proposed method achieves the best performance in garment state prediction within the simulation environment for all three types of garments. Specifically, our model achieves the prediction error of 2.94 cm, 2.84 cm and 3.16 cm, which significantly reduces the prediction error over other baselines. In the following garment-hanging experiment, we demonstrate the effectiveness of pre-hanging adjustment for garment-hanging. By using a model-based action sampling strategy, we can identify the optimal action to manipulate the garment to an ideal configuration for subsequent hanging. In the hanging experiments, our approach outperforms two end-to-end learning baselines by 12\%, 24\% and 10\%, respectively. Given that garments are highly deformable, a failure in the initial garment insertion can jeopardize the entire hanging process. Therefore, simply applying end-to-end imitation learning to such complex tasks is insufficient to ensure consistent success.

Our proposed method has also been validated in real-world experiments. From Table III, it can be observed that directly applying the simulation-based dynamics model may result in a significant sim-to-real gap, as simulated garments differ from real garments in size, appearance, texture, and mechanical properties. With our proposed residual model, \textbf{GraphGarment} reaches an average error of 3.15 cm, which only increases by 0.17 cm when compared with simulation results.


\section{Conclusions and Future Work}

In this paper, we propose \textbf{GraphGarment}, a GNN-based model incorporating a residual network for learning garment dynamics. It bridges the sim-to-real gap and enables a model-based action strategy for pre-hanging adjustment. Our approach is validated through a robotic bimanual hanging experiment in both simulation and the real-world. Experiment results also indicate the effectiveness of \textbf{GraphGarment} in prediction accuracy, sim-to-real transfer ability and task performance. 

However, our current system is based on the Markov property, where the next garment state is determined solely by the current state and action input. This could be improved by incorporating historical states into the prediction for greater accuracy. Additionally, future work could involve integrating advanced imitation learning algorithms, such as diffusion policies~\cite{chi2023diffusion}, with the garment dynamics model to enhance the manipulation of garment-related tasks.

\addtolength{\textheight}{-1cm}   


\bibliographystyle{IEEEtran}
\bibliography{main}   

\end{document}